\documentclass[twoside,leqno,twocolumn]{article}
\usepackage{ltexpprt}

\usepackage{amsfonts}
\usepackage{amsmath}
\usepackage{comment}
\usepackage{booktabs}

\makeatletter
\DeclareMathOperator*{\argmin}{arg\,min}

\makeatother

\usepackage[T1]{fontenc}
\usepackage[utf8]{inputenc}

\usepackage{times}
\usepackage{graphicx}
\usepackage{subfigure} 

\usepackage{algorithm}
\usepackage{algorithmic}

\begin{document}

\title{\Large Dynamic Stacked Generalization for Node Classification on Networks}
\author{Zhen Han\thanks{MaxPoint Interactive.} \\
\and
Alyson Wilson\thanks{Department of Statistics, North Carolina State University.}}
\date{}

\maketitle


\begin{abstract} \small\baselineskip=9pt We propose a novel stacked generalization (stacking) method as a dynamic ensemble technique using a pool of heterogeneous classifiers for node label classification on networks. The proposed method assigns component models a set of functional coefficients, which can vary smoothly with certain topological features of a node. Compared to the traditional stacking model, the proposed method can dynamically adjust the weights of individual models as we move across the graph and provide a more versatile and significantly more accurate stacking model for label prediction on a network. We demonstrate the benefits of the proposed model using both a simulation study and real data analysis.\end{abstract}

\section{Introduction}
\label{Introduction}
Network data and its relational structure have garnered tremendous attention in recent years. A network is composed of nodes and edges, where nodes represent interacting units and edges represent their relationships \cite{Goldenberg:2010:SSN:1734794.1734795}. 

Node classification or node labeling on a network is a problem where we observe labels on a subset of nodes and aim to predict the labels for the rest \cite{DBLP:journals/corr/abs-1101-3291}. There are various kinds of labels; for example, demographic labels such as age, gender, location, or social interests; labels such as political parties, research interests, or research affiliations. Collective inference estimates the labels of a set of related nodes simultaneously given a partially observed network by exploiting the relational auto-correlation of connected units
\cite{Jensen02linkageand}, and it has been demonstrated effective in reducing classification error for many applications \cite{Chakrabarti:1998:EHC:276304.276332, neville00iterative, DBLP:conf/kdd/JensenNG04, sen:aimag08}. 
Common collective inference methods are the Iterative Classification Algorithm (ICA) \cite{neville00iterative}, Gibbs Sampling (Gibbs), and Relaxation Labeling (RL) \cite{conf/icml/LuG03, Macskassy:2007:CND:1248659.1248693}.

In many cases, multiple types of relationships can be observed in the same network. For example, in a citation network, an edge can mean two papers have the same author, or they are published in the same journal, or one paper cites another. We may also observe additional node-level information, such as the title and abstract of a paper, which can potentially help increase the label classification accuracy. When multiple relations are present on a network, one can merge all the relations and sum the weights of common links to perform a typical collective classification \cite{Macskassy03asimple}. An alternative is to combine all of the information through an ensemble framework. Fürnkranz \cite{Furnkranz01hyperlinkensembles:} introduced hyperlink ensembles for classifying hypertext documents, where he suggests first predicting the label of each hyperlink attached to a document and then combining these individual predictions using ensembles to make a final prediction for the label of the target document. A different approach was proposed by Heß and Kushmerick \cite{Heb04iterativeensemble}, where they suggest training separate classifiers for the local and relational attributes and then combining the local and relational classifiers through voting. A local classifier is trained using only node-level, or local, features; for example, title, abstract, or year of publication. A relational classifier infers a node's label by using relational features; for example, the labels of the connected neighbors. Cataltepe et al.\ discussed a similar ensemble approach \cite{conf/mldm/CataltepeSBE11}, where they considered different voting methods, such as the weighted average, average, and maximum. Eldardiry and Neville \cite{Eldardiry11across-modelcollective} discussed an across-models collective classification method that formed ensembles of the estimates from multiple classifiers using a voting idea similar to collective inference to reduce variance. 

The above literature focuses on combining multiple classifiers through some type of aggregation. Preisach and Schmidt-Thieme \cite{Preisach:2008:ERC:1357641.1357645} proposed to use stacking instead of a simple voting as a more robust and powerful generalizing method to combine predictions made by local and relational classifiers. They suggest training each classifier independently and combining the predicted class probabilities from a pool of local and relational classifiers through stacking, which assigns constant weights to each classifier in a supervised fashion.  

Stacked generalization (stacking) \cite{Wolpert92stackedgeneralization} is a technique for combining multiple classifiers, each of which has been individually trained for a specific classification task, to achieve greater overall predictive accuracy. The method first trains individual classifiers using cross-validation on the training data. The original training data is called level-0 data, and the learned models are called level-0 classifiers. The prediction outcomes from the level-0 models are pooled for the second-stage learning, where a meta-classifier is trained. The pooled classification outcomes are called level-1 data and the meta-classifier is called the level-1 generalizer. 

Ting and Witten \cite{Ting97stackedgeneralization:} showed that for the task of classification, the best practice is to use the predicted class probabilities generated by level-0 models to construct level-1 data. 
Essentially, stacking learns a meta-classifier that assigns a set of weights to the class predictions made by individual classifiers. The traditional stacking model assumes the weight of each classifier is constant from instance to instance, which does not hold in general for many relational classifiers on a network. For example, the weighted-vote relational neighbor (wvRN) classifier \cite{Macskassy03asimple} infers a node's label by taking a weighted average of the class membership probabilities of its neighbors. One expects that its performance might be dependent on a node's topological characteristics in the graph; for example, the number of connected neighbors. On the other hand, local classifiers that are trained using only a node's local attributes are less dependent on its topological features. Consequently, when we combine local and relational classifiers, it is beneficial to have a set of weight functions instead of constant weights for each classifier. There has been some previous work on dynamically ensemble local and relational models \cite{conf/icml/McDowellA12, DBLP:conf/icdm/XiangN11}. However, they impose parametric models on the weight functions that are not flexible to capture complex weighting functions.

In this paper, we develop a dynamic stacking framework using a generalized varying coefficient model, which allows the weights for each classifier to vary smoothly with a node's topological characteristics in a non-parametric way. We illustrate the benefits of incorporating a node's topological features into stacking. To the best of our knowledge, this is the first work that considers non-parametric functional weight stacking. 

\section{Background and Motivation}
Network data can be represented by a graph $G = (V,E,Y)$ with vertices (nodes) $V$, edges (connections) $E = \{v_1,v_2\}, v_1,v_2 \in V$, and labels $Y$. The graph $G$ is partitioned into two sets of vertices, $V_{\text{train}}$ and $V_{\text{test}}$, with $V_{\text{train}} \cup V_{\text{test}} = V$ and $V_{\text{train}} \cap V_{\text{test}} = \emptyset$. We are given a classification problem with $C$ classes. Class labels, $y_i$, are observed for nodes in the training set $v_i \in V_{\text{train}}$, while the labels of the test set $V_{\text{test}}$ are unknown and need to be estimated. A relational classifier uses the attributes and/or labels from a node's connected neighbors to make predictions. However, unlike a typical classification problem, a node's neighbors may have missing attributes and/or labels, which in turn need to be estimated. Collective inference \cite{DBLP:conf/kdd/JensenNG04, sen:aimag08} has been developed to make joint inference on the test nodes and produce consistent results.
 
We examinine the Cora \cite{mccallum00automating} and the PubMed Diabetes \cite{sen:aimag08} data sets, where we evaluate the collective classification accuracy on nodes with various topological characteristics. We consider the wvRN classifier as the relational classifier \cite{Macskassy03asimple}, defined as follows, and the Iterative Classification Algorithm (ICA) \cite{conf/icml/LuG03, Macskassy:2007:CND:1248659.1248693} for collective inference as defined in Algorithm \ref{alg:ICA}.

\begin{Definition}{\rm For a given node $v_i \in V_{\text{test}}$, the wvRN classifier estimates the class probability  $P(y_i|\mathbb{N}_i)$ by the weighted average of the class membership probabilities in the neighborhood of $v_i$, $\mathbb{N}_i$:

\begin{equation}
P(y_i = c | \mathbb{N}_i) = \frac{1}{Z} \sum_{v_j \in \mathbb{N}_i} w_{i,j} P(y_j = c | \mathbb{N}_j),
\end{equation}
where $Z$ is a normalizing constant and $w_{i,j}$ is the weight associated with the edge between $v_i$ and $v_j$.}
\end{Definition}
Macskassy and Provost showed that the weighted-vote relational neighbor classifier is equivalent to the Gaussian-field model \cite{Macskassy:2007:CND:1248659.1248693}.

\begin{algorithm}[tb]
   \caption{Iterative Classification Algorithm (ICA)}
   \label{alg:ICA}
\begin{algorithmic}[1]
   \STATE For $v_i \in V_{test}$, initialize the node labels, $y_i$, with a dummy label \textbf{null}.
   \REPEAT
   \STATE Generate a random sequence of nodes, $O$, in $V_{\text{test}}$
   \FOR{node $v_i \in O$}
   \STATE Apply the relational classifier model, using only non-\textbf{null} labels from $\mathbb{N}_i$, the neighborhood of $v_i$, and output an estimated class membership probability vector. We ignore nodes that have not been classified, so if all labels in $\mathbb{N}_i$ are \textbf{null}, we assign the label \textbf{null} to $v_i$. 

    \STATE Assign the label, $c$, with the largest class membership probability to $v_i$. 

   \ENDFOR
   \UNTIL{class assignments for $V_{\text{test}}$ stop changing or a maximum number of iterations is reached.}
\end{algorithmic}
\end{algorithm}

The Cora data set is a public academic database composed of papers from Computer Science. It contains a citation graph with attributes/labels of each paper (including authors, title, abstract, book title, and topic labels). We only consider the topics of each paper as its label and ignore the other attributes. We remove papers with no topic labels and construct the data set by keeping the largest connected component in the network. The final data set is an unweighed and non-directional network, with 19,355 nodes and 58,494 edges. Labels are 70 topic categories, and each paper is classified into one of the categories. 

\begin{figure}[ht]
\vskip 0.2in
\begin{center}
\centerline{\includegraphics[width=\columnwidth]{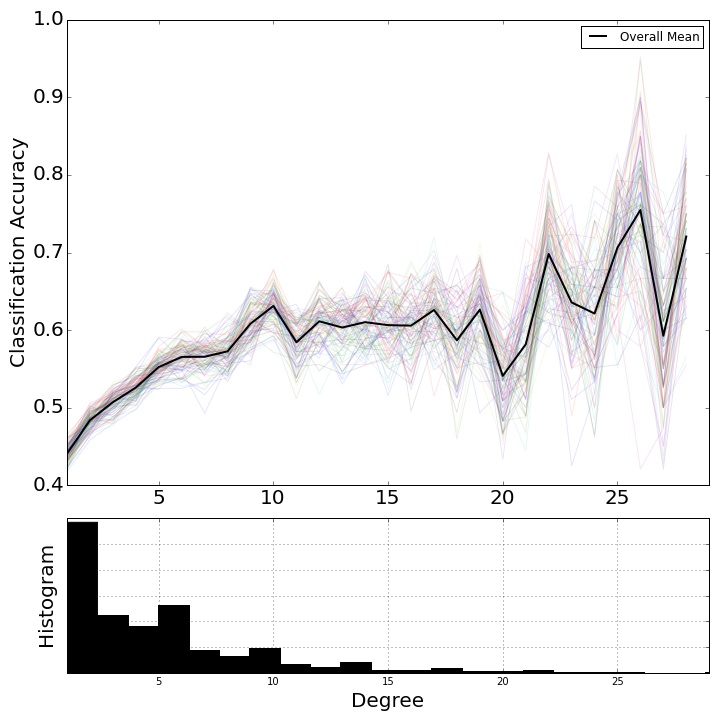}}
\caption{Classification accuracy for the relational classifier at different levels of node degree in the Cora data set.}
\label{fig:100_replication_90_testsize_degree_maximum}
\end{center}
\vskip -0.2in
\end{figure}

\begin{figure}[ht]
\vskip 0.2in
\begin{center}
\centerline{\includegraphics[width=\columnwidth]{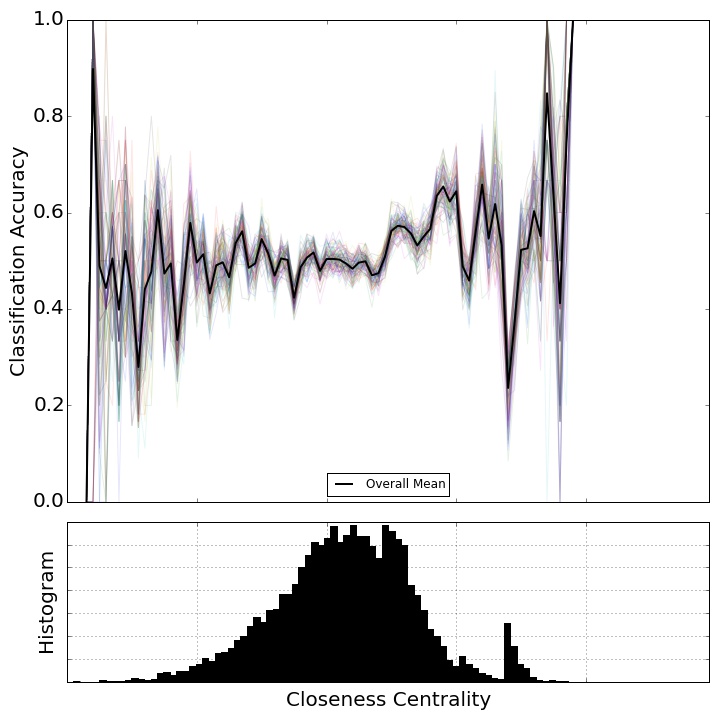}}
\caption{Classification accuracy for the relational classifier at different levels of node closeness centrality in the Cora data set.}
\label{fig:100_replication_90_testsize_closeness_centrality_maximum}
\end{center}
\vskip -0.2in
\end{figure}

We randomly sample 80\% of nodes from $V$ into $V_{test}$ and set their labels to \textbf{null}. We then make predictions using ICA on the nodes in $V_{test}$ and calculate the classification accuracy for different levels of degrees and closeness centrality. We repeat this experiment 100 times and the results are displayed in Figures (\ref{fig:100_replication_90_testsize_degree_maximum}) and (\ref{fig:100_replication_90_testsize_closeness_centrality_maximum}). In Figure (\ref{fig:100_replication_90_testsize_degree_maximum}), the classification accuracy of the wvRN classifier is dependent on the degree of $v_i$. As the count of a node's neighbor increases from 1 to 10, the average classification accuracy jumps from 45\% to more than 60\%. There are a limited number of nodes with degrees greater than 10, and thus the variance of the average classification accuracy goes up considerably. Closeness centrality is defined as the reciprocal of a node's total distance from all other nodes, which relates to the idea of ``being in the middle of things.'' Unlike degree, closeness centrality is a continuous variable. We binned its range into 100 equal-length intervals and calculated classification accuracy in each bin. From Figure (\ref{fig:100_replication_90_testsize_closeness_centrality_maximum}), we observe a steady upward trend in the classification accuracy near the center of the spectrum. There are not many nodes near the two ends of the spectrum, and this contributes to the large variation in the classification accuracy. 

\begin{figure}[ht]
\vskip 0.2in
\begin{center}
\centerline{\includegraphics[width=\columnwidth]{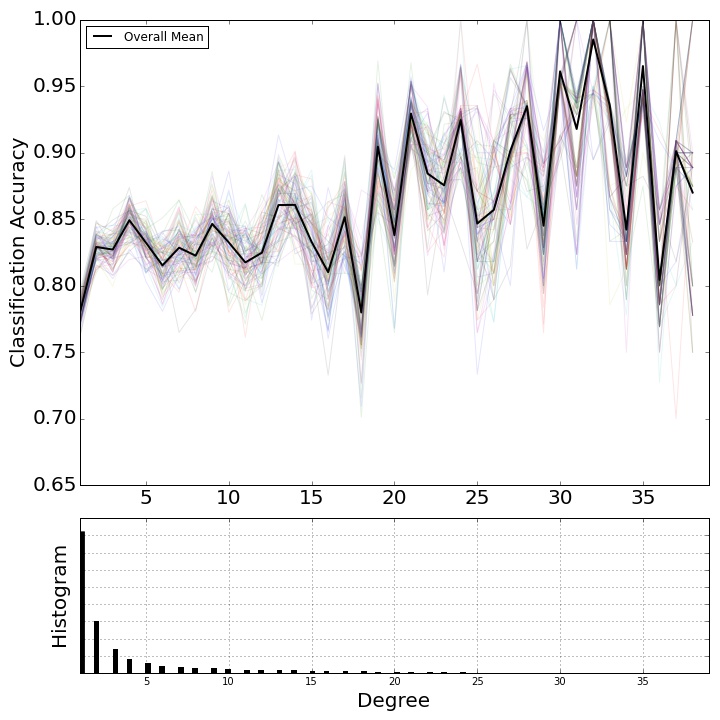}}
\caption{Classification accuracy for the relational classifier at different levels of node degree in the PubMed data set.}
\label{fig:Pubmed_100_replication_90_testsize_degree_maximum}
\end{center}
\vskip -0.2in
\end{figure}

\begin{figure}[ht]
\vskip 0.2in
\begin{center}
\centerline{\includegraphics[width=\columnwidth]{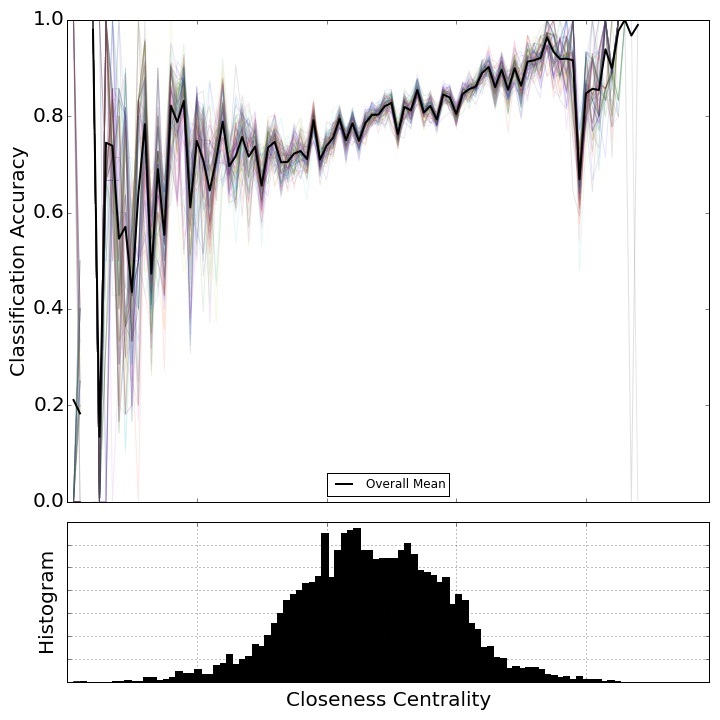}}
\caption{Classification accuracy for the relational classifier at different levels of node closeness centrality in the PubMed data set.}
\label{fig:Pubmed_100_replication_90_testsize_closeness_centrality_maximum}
\end{center}
\vskip -0.2in
\end{figure}

We performed the same analysis on the Pubmed Diabetes data set. The Pubmed data set is a medical database composed of diabetes-related medical papers derived from the PubMed database. The graph is a citation network with 19,717 papers and 44,338 edges. Each publication is assigned one of three categories as its label and a TF/IDF weighted word vector as an extra attribute. Here we ignore the extra attributes and only consider the topic category labels. We observe results similar to those from the Cora data set in Figures (\ref{fig:Pubmed_100_replication_90_testsize_degree_maximum}) and (\ref{fig:Pubmed_100_replication_90_testsize_closeness_centrality_maximum}).

\section{Dynamic Stacking Model}\label{sec:VaryingCoefficientStackingModel}
\subsection{Notation for the Stacked Generalization Model}\label{sec:GeneralStackingModel}
Stacked generalization (stacking) is a general method for combining multiple lower-level models to improve overall predictive accuracy \cite{Wolpert92stackedgeneralization, Ting97stackedgeneralization:}. Here we follow the notation in \cite{Ting97stackedgeneralization:}. Given data $\mathbb{D} = \{ (y_i,\boldsymbol{x}_i)$ for $i = 1, \cdots, N \}$, let $\boldsymbol{x}_i$ be the feature vector and $y_i$ the label of the $i$-th observation. Here we focus on categorical responses for $y$, and assume $y$ has $C$ categories. We first randomly partition the data into $J$ roughly equal-sized parts $\mathbb{D}_1, \mathbb{D}_2, \cdots, \mathbb{D}_J$. Define $\mathbb{D}_j$ and $\mathbb{D}_{-j} = \mathbb{D} - \mathbb{D}_j$ to be the test and training data sets for the $j$-th fold of a $J$-fold cross validation. 

Suppose we have $K$ classifiers. We train each of the $K$ classifiers using the training set $\mathbb{D}_{-j}$ with results $\mathbb{M}_{k}$. $\mathbb{M}_{1}, \cdots, \mathbb{M}_{K}$ are called \textit{level-0 models}. We then apply the $K$ classifiers on the test set $\mathbb{D}_j$ and denote $\boldsymbol{z}_{ik} = \mathbb{M}_{k}(\boldsymbol{x}_i), \boldsymbol{x}_i \in \mathbb{D}_j$ as the estimated class probability vector from $\mathbb{M}_{k}$ for $\boldsymbol{x}_i \in D_j$. We repeat this process for $j = 1, \cdots, J$ and collect the outputs from the $K$ models to form the \textit{level-1 data} as follows:
\begin{equation}
D_{\text{level 1}} = \{(y_i,\boldsymbol{z}_{i1},\cdots,\boldsymbol{z}_{iK}), \text{ for } i = 1 \cdots, N \}.
\end{equation}
$\boldsymbol{z}_{ik}$ is the predicted class probability vector from $\mathbb{M}_{K}$ for observation $i$, and therefore $\sum_c z_{ikc} = 1$. We drop the last element $z_{ikC}$ from vector $\boldsymbol{z}_{ik}$ to avoid multicollinearity issues. We then fit a supervised classification model, $\tilde{\mathbb{M}}$, using the level-1 data, which is called the \textit{level-1 generalizer}. 

For prediction, a new observation $\boldsymbol{x}_{\text{new}}$ is input into the $K$ low-level classifiers, $\mathbb{M}_{1}, \cdots,\mathbb{M}_{K}$. The estimated class probability vectors, $\boldsymbol{z}_1, \cdots, \boldsymbol{z}_K$, are then concatenated and input into $\tilde{\mathbb{M}}$, which outputs the final class estimate for that observation.

For classification on networks, Preisach and Schmidt-Thieme \cite{Preisach:2008:ERC:1357641.1357645} adapted the stacking technique and combined a local classifier with a relational classifier using a logistic regression model as the level-1 generalizer. However, in their paper, the coefficients in the logistic regression are constant, meaning the weights of individual component classifier are ``static'' from node to node. From previous observations in Figures (\ref{fig:100_replication_90_testsize_degree_maximum}), (\ref{fig:100_replication_90_testsize_closeness_centrality_maximum}), (\ref{fig:Pubmed_100_replication_90_testsize_degree_maximum}), and (\ref{fig:Pubmed_100_replication_90_testsize_closeness_centrality_maximum}), the accuracy of a relational classifier is often dependent on some topological feature of a node. Therefore, it could be beneficial to ``dynamically'' allocate the weights of individual classifiers based on some other variable. We discuss a dynamic stacking model using a generalized varying coefficient model in the next section to account for this observation. Compared with competing methods that also consider dynamic stacking \cite{conf/icml/McDowellA12, DBLP:conf/icdm/XiangN11}, the proposed model is non-parametric, more flexible, and can learn more complex weighting functions.

\subsection{Generalized Varying Coefficient Model through Smoothing Splines}
Here we develop a dynamic stacking model using a generalized varying coefficient model. Instead of having a set of constant coefficients in the regression, we allow the coefficients to vary as smooth functions of other variables. The generalized varying-coefficient model was proposed by Hastie and Tibshirani \cite{HastieTibshirani1993} and was reviewed in \cite{VaryingCoefficientModels08Fan}. 

Similar to traditional stacked generalization, the inputs for the dynamic stacking model are the assembled outputs from multiple level-1 classifiers, along with an extra covariate: $\{(y_i, \boldsymbol{Z}_i, u_i), \text{ for } i = 1, \cdots, N \}$. $y_i$ is the true class label of an observation, $\boldsymbol{Z}_i$ is the concatenated predicted class membership vector from a pool of inhomogeneous classifiers with dimension $p$. Each of the component classifiers could potentially look at a different set of features of an instance and make a prediction from its point of view. $u_i$ is an ``extra'' covariate of a observation, which presumably would affect the prediction accuracy made by at least one classifier. Here we focus on the case where $y_i$ is binary and $u_i$ is continuous. One can easily extend this method to multi-class classification problems by using a one-vs-all strategy.

The regression function is modeled as:
\begin{align}\label{eq:regressionModel}
g(m(U_i,\boldsymbol{Z}_i)) = & \beta_0 + \boldsymbol{Z}_i^T \boldsymbol{\beta}(U_i) \\
 = & \beta_0 + \sum^p_{j=1} Z_{ij} \beta_j(U_i) \nonumber 
\end{align}
where $g(\cdot)$ is the logit link function, $\boldsymbol{\beta}(\cdot)$ is the functional coefficient vector that varies smoothly with an extra scalar covariate, and $\beta_0$ is a constant intercept. Instead of a constant intercept, one can trivially add a functional intercept by appending $1$ to $\boldmath{Z_i}$. However, in this paper, we focus on the constant intercept case. We assume that each functional coefficient $\beta_j(\cdot) \text{ for } j = 1, \cdots, p$ can be approximated by spline functions:
\begin{align}\label{eq:betaSpline}
\beta_j(\cdot) = \sum_{k=1}^{K_j} \eta_{jk} B_{jk}(\cdot), \text{ for } j = 1, \cdots, p,
\end{align}
where for each $\beta_j$, $B_{jk}(\cdot) \text{ for } k = 1,  \cdots, K_j$ is a set of spline basis functions. Without loss of generality, we use the same set of B-spline basis functions for all $\beta_1(\cdot), \cdots, \beta_p(\cdot)$. Henceforth, we denote the set of B-spline basis functions as $B_1(\cdot), \cdots, B_K(\cdot)$, where $K$ is the number of basis functions. We can then rewrite equation (\ref{eq:betaSpline}) as: 
\begin{align}\label{eq:betaSpline2}
\beta_j(\cdot) = \sum_{k=1}^{K} \eta_{jk} B_{k}(\cdot) \text{ for } j = 1, \cdots, p.
\end{align}
We substitute equation (\ref{eq:betaSpline2}) into equation (\ref{eq:regressionModel}) and rewrite the regression function as
\begin{align}
g(m(U_i,\boldsymbol{Z_i}))  & = \beta_0 + \sum_{j=1}^p \sum_{k=1}^K \boldsymbol{Z}_{ij} \eta_{jk} B_k(U_i) \\
& = \beta_0 + \boldsymbol{Z}_i^T \boldsymbol{B}(U_i) \boldsymbol{\eta}, \nonumber
\end{align}

Denote $\boldsymbol{B}_{*}(U) = (B_{1}(U), \cdots, B_{K}(U))_{1 \times K} $. We can express $\boldsymbol{B}(U)$ as:
\[\boldsymbol{B}(U) = 
\begin{bmatrix}
    \boldsymbol{B}_{*}(U) & \dots & 0 \\
    \vdots & \ddots & \vdots \\
    0 & \cdots & \boldsymbol{B}_{*}(U) \\
\end{bmatrix}_{p \times pK}
\]
and express $\boldsymbol{\eta}$ as:
\[\boldsymbol{\eta} = 
\begin{bmatrix}
    \eta_{11} \\
    \vdots \\
    \eta_{pK} \\
\end{bmatrix}_{pK}
\].

We can estimate $\beta_0, \beta_1(\cdot), \cdots, \beta_p(\cdot)$ by directly minimizing:
\begin{align}\label{eq:objective}
\hat{\beta_0}, \hat{\beta_1}(\cdot), \cdots, \hat{\beta_p}(\cdot) &= \argmin_{\beta_0,\beta_1(\cdot), \cdots, \beta_p(\cdot)} \\
&-\sum_{i=1}^N \ell(\beta_0 + \sum_{j=1}^p Z_{ij} \beta_j(U_i),y_i) \nonumber \\
&+ \lambda \sum_{j=1}^p \int (\beta_j^{''}(x))^2 dx \nonumber
\end{align}
where $\ell(g(m(U_i,\boldsymbol{Z}_i)),y_i)$ is the log-likelihood function of the logistic regression, $\lambda \sum_{j=1}^p \int (\beta_j^{''}(x))^2 dx$ is a smoothness penalty term that controls the total curvature of the fitted $\beta_j(\cdot)$ for $j = 1, \cdots, p$, and $\lambda$ is a
a smoothing parameter that controls the trade-off between model fit and the roughness of the fitted $\beta_j(\cdot)$s. When $\lambda \rightarrow 0$, we have a set of wiggly $\beta_j(\cdot)$s; as $\lambda \rightarrow \infty$, the minimization of  (\ref{eq:objective}) will produce a set of linear $\beta_j(\cdot)$s. 

For the constant intercept case, one can absorb $\beta_0$ into $\boldsymbol{\eta}$ as:
\[\boldsymbol{\eta}^* = 
\begin{bmatrix}
    \beta_0 \\
    \eta_{11} \\
    \hdotsfor{1} \\
    \eta_{pK} \\
\end{bmatrix}_{1 + pK}
\]
and append a constant $1$ to the beginning of the product, $\boldsymbol{Z}_i^T \boldsymbol{B}(U_i)$. We can write the optimization in equation (\ref{eq:objective}) w.r.t.\ $\boldsymbol{\eta}^*$ as:
\begin{align}\label{eq:eta_objective}
\hat{\boldsymbol{\eta}^*} = \argmin_{\boldsymbol{\eta}^*} \{ -\ell(\boldsymbol{\eta}^*) + \lambda \, \boldsymbol{\eta}^{*T} \boldsymbol{H} \boldsymbol{\eta}^* \},
\end{align}
where $\boldsymbol{H}$ is the assembled penalty matrix:
\[\boldsymbol{H} = 
\begin{bmatrix}
    0 & \hdotsfor{2} &0 \\
    \hdotsfor{1} & H^1 & 0 & 0 \\
    \hdotsfor{1} & \hdotsfor{1} & H^j & \hdotsfor{1}\\
    0 & 0 & 0 & H^p \\
\end{bmatrix}_{(1 + pK) \times (1 + pK)} 
\]
$H^j$ is the smoothness penalty matrix for $\beta_j(\cdot)$, and $\{ H^j \}_{mn} = \int B^{''}_m(x)B^{''}_n(x) dx$ for $m, n = 1, \cdots, K$. Since we are using the same set of basis functions, $H^1 = \cdots = H^p$. It can be shown that $-\ell(\boldsymbol{\eta}^*)$ is convex w.r.t.\ $\boldsymbol{\eta}^*$. Also, one can show that $\boldsymbol{H}$ is positive semi-definite, so $\lambda \, \boldsymbol{\eta}^{*T} \boldsymbol{H} \boldsymbol{\eta}^*$ is convex w.r.t.\ $\boldsymbol{\eta}^*$ as well. Therefore, there exists a unique $\hat{\boldsymbol{\eta}^*}$ that optimizes equation (\ref{eq:eta_objective}). 

Given a specified smoothness penalty parameter $\lambda$, to estimate $\boldsymbol{\eta}^*$, we employ an iterative Newton-type optimization method by directly calculating the derivatives of the objective function in equation  (\ref{eq:eta_objective}). The smoothness penalty parameter $\lambda$ can be chosen by cross-validation, where, for a range of $\lambda$ values, we iteratively leave out a subset of the training data, fit the model using the rest of the data, and compute the prediction error on the held out data set. The best $\lambda$ is set to the one with the smallest objective function value.

\section{Simulation Study}
Here we compare the performance of the dynamic stacking method against standard benchmarks using simulated data sets.  \cite{Preisach:2008:ERC:1357641.1357645} used a standard logistic regression model as the level-1 generalizer to combine a local and a relational classifier. \cite{RegularizedStacking} suggested that regularization is necessary to reduce over-fitting and increase predictive accuracy, and they considered lasso regression, ridge regression, and elastic net regression. In our simulation study, we use lasso regression, ridge regression, and logistic regression as benchmark level-1 generalizers, and for each of the benchmark generalizers, we experiment with adding an additional covariate and/or interaction terms into the stacking, and compare their performance with the dynamic stacking model. 

For the simulation, $N = 2000$ observations are generated for $i = 1, \cdots, N$. $Z_{1i}$ and $Z_{2i}$ are the predicted positive class probabilities from two classifiers, $\mathbb{Z}_1$ and $\mathbb{Z}_2$, and they are generated independently from a uniform distribution on $[0,1]$. $w_i$ is the error term which follows a normal distribution, $N(0,1)$. $u_i$ is an extra covariate which may affect the weight of a classifier, and it is generated from a uniform distribution on $[0,1]$. Finally, the response $y_i$ is generated from a Bernoulli distribution with $p(y_i = 1)$ specified as one of the following three cases. In case 1, the classifier weights are not dependent on $u$, while case 2 has linear dependence, and case 3 has non-linear dependence.\\

\noindent Case 1: 
\begin{equation*}\label{eq:Model1}
\text{logit}(p(y_i = 1)) =  -3 + 3Z_{1i} + 3Z_{2i} + w_i 
\end{equation*}

\noindent Case 2:
\begin{equation*}\label{eq:Model2}
\text{logit}(p(y_i = 1)) =  -3 + 3 u_i Z_{1i} + 3Z_{2i} + w_i 
\end{equation*}

\noindent Case 3:
\begin{equation*}\label{eq:Model3}
\text{logit}(p(y_i = 1)) =  -3 + 3 \sin(6u_i) Z_{1i} + 3Z_{2i} + w_i
\end{equation*}
For training and evaluation, the $N$ observations are evenly split into training and testing sets. We train the dynamic stacking model and benchmark methods using the training set, where the penalty parameters of the proposed method and benchmarks are selected by 10-fold cross-validation. The fitted models are then applied to predict on the testing set, and the final prediction accuracy on the test set is measured by the Area Under the Curve (AUC) from prediction scores as shown in Table (\ref{tab:simulationresult}). For methods$^1$ (Logistic$^1$, Lasso$^1$, and Ridge$^1$), the inputs to the level-1 generalizer are $\{(y_i, Z_{1i},Z_{2i})\}$. For methods$^2$, we add the additional covariate $u$ into the input: $\{(y_i, Z_{1i},Z_{2i}, u_i)\}$. For methods$^3$, in addition to $u$, we further add its interaction with $Z_{1i},Z_{2i}$ into the input: $\{(y_i, Z_{1i},Z_{2i},u_i,Z_{1i}u_i, Z_{2i}u_i)\}$. 

\begin{table}[t]
\caption{AUC score comparison between the proposed method and benchmarks. For level-1 generalizers, methods$^1$ use $Z_{1i}$ and $Z_{2i}$ as covariates, methods$^2$ contains $U_i$ as an extra feature, and methods$^3$ further include linear interaction terms $U_i Z_{1i}$, and $U_i Z_{2i}$. The standard deviation of the accuracy score is calculated from 50 repetitions and is shown in parenthesis.}
\label{tab:simulationresult}
\vskip 0.15in
\begin{center}
\begin{small}
\begin{sc}
\begin{tabular}{p{2.8cm}p{1cm}p{1cm}p{1cm}r}
\hline
\toprule
Methods & Case 1 & Case 2 & Case 3 \\
\midrule
Random Guess & 0.49 (0.02) & 0.50 (0.02) & 0.51 (0.02) \\
$Z_1$ Only & 0.68 (0.02) & 0.60 (0.02) & 0.54 (0.02) \\
$Z_2$ Only &  0.68 (0.02) & 0.68 (0.02) &  0.67 (0.02) \\
Logistic$^1$ & 0.75 (0.02) & 0.71 (0.02) & 0.67 (0.01) \\
Lasso$^1$ &  0.75 (0.02) &  0.71 (0.02) &  0.67 (0.01) \\
Ridge$^1$               &  0.75 (0.02) &  0.71 (0.02) &  0.67 (0.01) \\
Logistic$^2$ &  0.75 (0.02) &  0.73 (0.02) &  0.75 (0.02) \\
Lasso$^2$               &  0.75 (0.02) &  0.72 (0.02) &  0.74 (0.02) \\
Ridge$^2$               &  0.75 (0.02) &  0.72 (0.02) &  0.74 (0.02) \\
Logistic$^3$ &  0.75 (0.01) &  0.73 (0.02) &  0.76 (0.01) \\
Lasso$^3$               &  0.74 (0.02) &  0.73 (0.02) &  0.76 (0.01) \\
Ridge$^3$               &  0.74 (0.02) &  0.73 (0.02) &  0.76 (0.01) \\
\bottomrule
Proposed Method      &  \textbf{0.75 (0.02)} &  \textbf{0.73 (0.02)} &  \textbf{0.79 (0.01)} \\
\hline
\end{tabular}
\end{sc}
\end{small}
\end{center}
\vskip -0.1in
\end{table} 

From Table (\ref{tab:simulationresult}), the dynamic stacking model performs no worse than the standard methods under all scenarios. It has better performance than the ``static'' stacking models when there is an underlying non-linear dependency between a classifier's performance and the extra covariate. In case 1, the dynamic stacking model generalizes to the traditional stacking models and does not over fit the data. In case 2, where a linear dependency exists, the proposed model generalizes to methods$^3$, where interaction terms with the extra covariate are added into the ``static'' stacking model. In case 3, where a non-linear dependency exists, the dynamic stacking model outperforms all benchmarks.

\begin{table*}[t]
\centering
\caption{Mean accuracy score on 100 randomized experiments using Cora and PubMed data set.}
\label{model-comparison1}
\begin{tabular}{@{}ccccccc@{}}
\toprule
       & Xiang    & McDowell & Lasso    & Ridge    & Logistic & Proposed          \\ \midrule
Cora   & 0.917476 & 0.918087 & 0.917898 & 0.918106 & 0.918401 & \textbf{0.919270} \\
PubMed & 0.879380 & 0.879205 & 0.889090 & 0.889125 & 0.889213 & \textbf{0.891499} \\ \bottomrule
\end{tabular}
\end{table*}

\begin{table*}[t]
\centering
\caption{Pairwise accuracy score comparison between the proposed method versus each competing method in 100 randomized experiments. The values shown are the mean accuracy difference between the proposed method and each method and the p-value for $H_0:$ the proposed method is less accurate than that benchmark method. }
\label{model-comparison2}
\begin{tabular}{@{}cccccc@{}}
\toprule
Proposed       & vs Xiang          & vs McDowell       & vs Lasso          & vs Ridge          & vs Logistic       \\ \midrule
Cora   & 0.1794                    & 0.1183                    & 0.1372                    & 0.1164                    & 0.0869                    \\
       & \textbf{p \textless 0.01} & \textbf{p \textless 0.01} & \textbf{p \textless 0.01} & \textbf{p \textless 0.01} & \textbf{p \textless 0.01} \\
PubMed & 1.2119                    & 1.2294                    & 0.2409                    & 0.2374                    & 0.2286                    \\
       & \textbf{p \textless 0.01} & \textbf{p \textless 0.01} & \textbf{p \textless 0.01} & \textbf{p \textless 0.01} & \textbf{p \textless 0.01} \\ \bottomrule
\end{tabular}
\end{table*}

\section{Real Data Analysis}
Here we revisit the Cora data set \cite{mccallum00automating}, where we use paper titles as node attributes and topic classification as labels. We remove nodes with no title or topic classifications, and the final graph contains 11,187 nodes and 33,777 edges. Seventy topic categories are used as labels, and each paper belongs to one of the categories. For simplicity, we convert the classification problem into a binary classification problem by giving a positive label if the topic falls under the \verb|/Artificial_Intelligence/| category. We then use the closeness centrality of each node in the graph as an additional covariate in stacking. 

\begin{figure}[ht]
\vskip 0.2in
\begin{center}
\centerline{\includegraphics[width=\columnwidth]{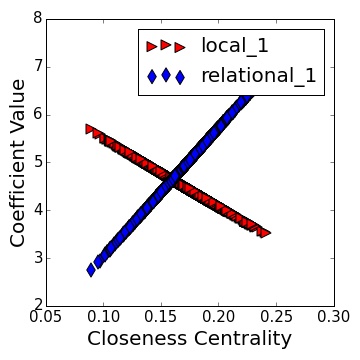}}
\caption{One set of fitted coefficient curves.}
\label{fig:one-curve}
\end{center}
\vskip -0.2in
\end{figure}

We split all the nodes on the graph into a 20\% training set and an 80\% testing set. On the training set, we observe the titles and the topic classification label of each paper, while on the test set, we only observe the titles. We fit a local classifier using the word vector representation of its title only (Naive Bayes), and a relational classifier (ICA + wvRN) using only the labels from a paper's neighbor. We then fit a dynamic stacking model using the output from the two classifiers with their coefficients being smooth functions of the closeness centrality of a node. The smoothness penalty parameter is chosen by 10-fold cross-validation. One set of fitted coefficient curves for the two classifiers are shown in Figure \ref{fig:one-curve}. It allocates a higher weight on the relational classifier when a node has a high closeness centrality value and relies on the local classifier for nodes with a small closeness centrality value. This mirrors our observations from the previous discussion. 

\begin{figure*}[ht]
\vskip 0.2in
\begin{center}
  \subfigure[]{\includegraphics[width=0.24\linewidth]{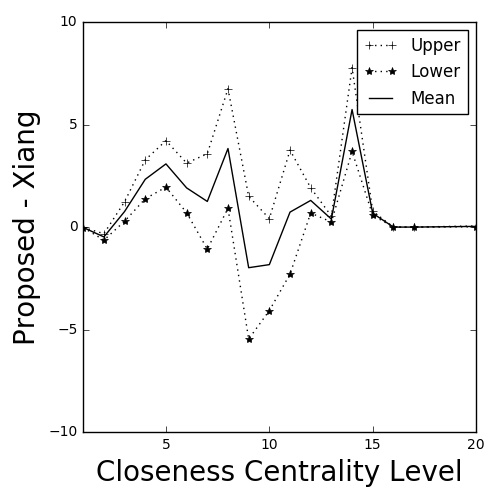}}\quad
  \subfigure[]{\includegraphics[width=0.24\linewidth]{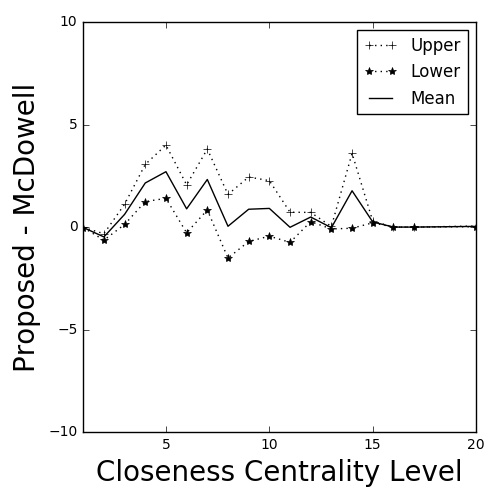}} 
    \subfigure[]{\includegraphics[width=0.24\linewidth]{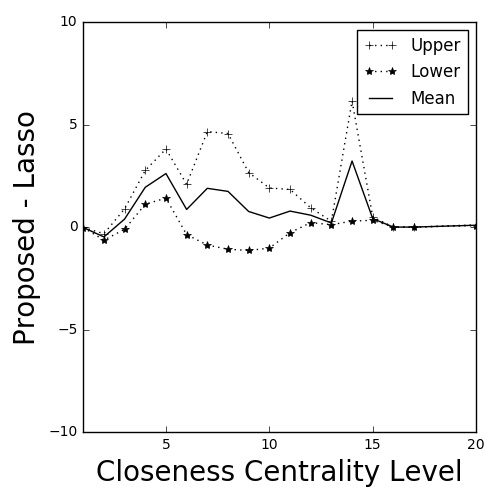}} \\
      \subfigure[]{\includegraphics[width=0.24\linewidth]{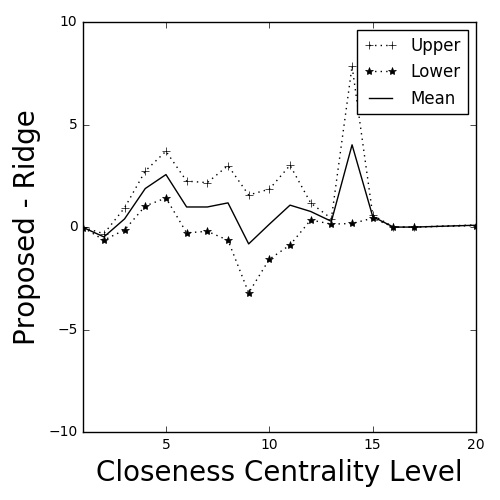}} 
    \subfigure[]{\includegraphics[width=0.24\linewidth]{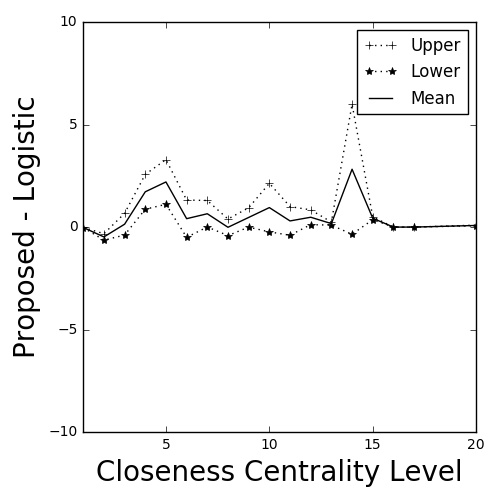}}
      \subfigure[]{\includegraphics[width=0.24\linewidth]{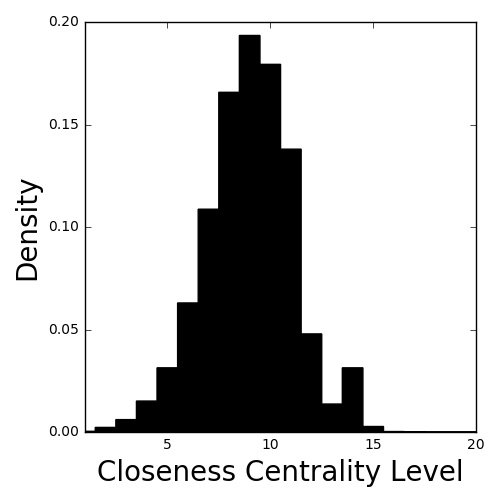}}
\caption{For each of the 100 repetitions, we calculate the difference in the number of correctly classified nodes at different closeness centrality levels between the dynamic stacking model and benchmarks. In (a) -- (e), we calculate the group mean and a 95\% confidence interval. (f) shows a density distribution of the closeness centrality for all nodes in the graph.}
\end{center}
\label{fig:DigitizeClosenessComparison}
\vskip -0.2in
\end{figure*}

We compared the dynamic stacking model with the traditional stacking model on multiple standard level-1 generalizers (lasso, ridge, and logistic regression), all of which ignore the closeness centrality of a node during stacking. The penalty parameters for lasso and ridge regression are chosen by 10-fold cross-validation. We also implemented ensemble classification methods from \cite{DBLP:conf/icdm/XiangN11,conf/icml/McDowellA12}. In \cite{DBLP:conf/icdm/XiangN11}, Xiang and Neville proposed a parametric weighting scheme to combine a local classifier and a relational classifier where model parameters are chosen by cross-validation. In \cite{conf/icml/McDowellA12}, outputs from local and relational classifier are combined thorough a concept of label regularization and the model they proposed has no additional parameters. We repeat the train-test process 100 times randomly and record the accuracy score for each run. 

We also performed model comparison using the PubMed data with the same general setup. Input to the local classifier is the TF/IDF representation of a paper, and we fit a dynamic stacking model using the output from the local classifier and the relational classifier with their coefficients being smooth functions of the degree centrality of a node. The classification accuracy comparison between the proposed method and the benchmarks is shown in Table (\ref{model-comparison1}) and Table (\ref{model-comparison2}), where  the accuracy is defined as $\sum^{N}_{i = 1} \mathbb{I}_{y_i = \hat{y}_i} / N$ where $\hat{y}_i = \mathbb{I}_{\hat{p}_i > 0.5}$.

By assuming the normality of the classification accuracy difference distribution, the dynamic stacking mode outperforms all benchmarks at p-value $< 0.01$. For the Cora data set, Figure 6 shows the source of the accuracy improvement. For each of the 100 repetitions, we calculate the difference in the absolute number of correctly classified nodes at different closeness centrality levels between the proposed model and benchmarks. The dynamic stacking model outperforms the benchmarks near the two ends of the closeness centrality spectrum where the balance between the local and relational classifier shifts considerably. For the majority of nodes in this data set, their closeness centrality clusters tightly around a specific value, which leaves little room for the dynamic stacking model to improve much beyond its static-weight counterparts in terms of the overall accuracy. However, for the nodes that are near the two extremes of the closeness centrality spectrum, we do see a significant improvement by using the dynamic stacking method.  

\section{Discussion}
In this paper, by examining two public data sets, Cora and PubMed, we illustrate the motivation for incorporating node topological characteristics into stacked generalization for node classification on networks. We then develop a novel dynamic stacking method with functional weights for component models, each of which can vary smoothly with an extra covariate. Simulation studies show that the proposed method generalizes well to the benchmarks when the data does not present complex patterns, and outperforms all benchmarks otherwise. Real data analysis using Cora and PubMed shows the proposed method has a small yet significant improvement on the classification accuracy compared with traditional stacking models. Further analysis shows that most of the accuracy improvement comes from nodes near the two extremes of the closeness centrality spectrum where the balance between the local and relational classifier shifts the most. The limited number of nodes in that region explains the small improvement on the overall classification accuracy. However, for the nodes near the extremes of closeness centrality, we do see a considerable improvement on the classification accuracy using the dynamic stacking method.     
Overall, the dynamic stacking model allows the composition of the stacking model to change as we move across the network, and thus it potentially provides a more versatile and accurate stacking model for label prediction on a network.

\section{Future Work}
The proposed dynamic stacking model is a direct extension of the traditional stacking model, which uses logistic regression as level-1 generalizer. As discussed in \cite{RegularizedStacking}, this model tends to overfit, especially when combining a large pool of noisy classifiers. To mitigate this problem, one can add a group-lasso penalty over the model coefficients, $\boldsymbol{\eta}^*$, into equation (\ref{eq:eta_objective}). Coefficients can be naturally grouped if they are the basis coefficients for the same $\beta_j(\cdot)$: \{($\eta_{j1}, \cdots, \eta_{jK}), \text{ for } j = 1, \cdots, p \}$. By adding a group-lasso penalty, the dynamic stacking model is more robust to the noise in the level-1 generalizing process. 

\section{Acknowledgment}
This material is based upon work supported in part with funding from the Laboratory for Analytic Sciences (LAS). Any opinions, findings, conclusions, or recommendations expressed in this material are those of the author(s) and do not necessarily reflect the views of the LAS and/or any agency or entity of the United States Government.

\bibliography{Reference}
\bibliographystyle{plain}

\end{document}